# Translation of Pronominal Anaphora between English and Spanish: Discrepancies and Evaluation

**Jesús Peral** JPERAL@DLSI.UA.ES
*Dept. Lenguajes y Sistemas Informáticos*
*University of Alicante*
*Alicante, SPAIN*

**Antonio Ferrández** ANTONIO@DLSI.UA.ES
*Dept. Lenguajes y Sistemas Informáticos*
*University of Alicante*
*Alicante, SPAIN*

## Abstract

This paper evaluates the different tasks carried out in the translation of pronominal anaphora in a machine translation (MT) system. The MT interlingua approach named AGIR (*Anaphora Generation with an Interlingua Representation*) improves upon other proposals presented to date because it is able to translate intersentential anaphors, detect co-reference chains, and translate Spanish zero pronouns into English—issues hardly considered by other systems. The paper presents the resolution and evaluation of these anaphora problems in AGIR with the use of different kinds of knowledge (lexical, morphological, syntactic, and semantic). The translation of English and Spanish anaphoric third-person personal pronouns (including Spanish zero pronouns) into the target language has been evaluated on unrestricted corpora. We have obtained a precision of 80.4% and 84.8% in the translation of Spanish and English pronouns, respectively. Although we have only studied the Spanish and English languages, our approach can be easily extended to other languages such as Portuguese, Italian, or Japanese.

## 1. Introduction

The anaphora phenomenon can be considered one of the most difficult problems in natural language processing (NLP). The etymology of the term *anaphora* originates with the Ancient Greek word "anaphora" ($\alpha\nu\alpha\varphi o\rho\alpha$), which is made up of the separate words, $\alpha\nu\alpha$ ("back, upstream, back in an upward direction") and $\varphi o\rho\alpha$ ("the act of carrying"), and which denotes the act of carrying back upstream.

Presently, various definitions of the term *anaphora* exist, but the same concept underlies all of them. Halliday & Hassan (1976) defined anaphora as "the cohesion (presupposition) which points back to some previous item." A more formal definition was proposed by Hirst (1981), which defined anaphora as "a device for making an abbreviated reference (containing fewer bits of disambiguating information, rather than being lexically or phonetically shorter) to some entity (or entities) in the expectation that the receiver of the discourse will be able to disabbreviate the reference and, thereby, determine the identity of the entity." Hirst refers to the entity as an *anaphor*, and the entity to which it refers is its *antecedent*:

- *[Mary]$_i$* went to the cinema on Thursday. *She$_i$* didn't like the film.

In this example, the pronoun *she* is the anaphor and the noun phrase *Mary* is the antecedent. This type of anaphora is the most common type, the so-called *pronominal anaphora*.

The anaphora phenomenon can be further broken down into two processes: that of resolution and generation. "Resolution" refers to the process of determining the antecedent of an anaphor; "generation" is the process of creating references over a discourse entity.

In the context of machine translation, the resolution of anaphoric expressions is of crucial importance in order to translate/generate them correctly into the target language (Mitkov & Schmidt, 1998). Solving the anaphora and extracting the antecedent are key issues for correct translation into the target language. For instance, when translating into languages which mark the gender of pronouns, resolution of the anaphoric relation is essential. Unfortunately, the majority of MT systems do not deal with anaphora resolution, and their successful operation usually does not go beyond the sentence level.

We have employed a computational system that focuses on anaphora resolution in order to improve MT quality and have then measured the improvements. The SUPAR (*Slot Unification Parser for Anaphora Resolution*) system is presented in the work of Ferrández, Palomar, & Moreno (1999). This system can deal with several kinds of anaphora with good results. For example, the system resolves pronominal anaphora in Spanish with a precision rate of 76.8% (Palomar et al., 2001); it resolves one-anaphora in Spanish dialogues with a precision rate of 81.5% (Palomar & Martínez-Barco, 2001), and it resolves definite descriptions in Spanish direct anaphora and bridging references with precision rates of 83.4% and 63.3%, respectively (Muñoz, Palomar, & Ferrández, 2000). In the work presented here, we have used an MT system exclusively for pronominal anaphora resolution and translation. This kind of anaphora is not usually taken into account by most of the MT systems, and therefore pronouns are usually translated incorrectly into the target language. Although we have focused on pronominal anaphora, our approach can be easily extended to other kinds of anaphora, such as one-anaphora or definite descriptions previously resolved by the SUPAR system.

It is important to emphasize that in this work we only resolve and translate personal pronouns in the third person whose antecedents appear before the anaphor—that is, an anaphoric relation between the pronoun and the antecedent is established, and cataphoric relations (in which the antecedent appears after the anaphor) are not taken into account.

This paper focuses on the evaluation of the different tasks carried out in our approach that lead to the final task: the translation of the pronominal anaphora into the target language. The main contributions of this work are a presentation and evaluation of the multilingual anaphora resolution module (English and Spanish) and an exhaustive evaluation of the pronominal anaphora translation between these languages.

The paper is organized as follows: Section 2 shows the anaphora-resolution needs in MT and the deficiencies of traditional MT systems to resolve this phenomenon conveniently. Section 3 presents the analysis module of our approach. In Section 4, we identify and evaluate the NLP problems related to pronominal anaphora resolved in our system. Section 5 presents the generation module of the system. In Section 6, the generation module is evaluated in order to measure the efficiency of our proposal. Finally, we present our conclusions.

## 2. Anaphora Resolution and its Importance in MT

As noted earlier, anaphora resolution is of crucial importance in order to translate anaphoric expressions correctly into a target language. Let us consider the sentences (Hutchins & Somers, 1992):

1. *[The monkey]$_i$* ate the banana because *it$_i$* was hungry.
2. The monkey ate *[the banana]$_i$* because *it$_i$* was ripe.
3. The monkey ate the banana because *it* was tea-time.

In each sentence the pronoun *it* refers to something different: in sentence (1), it refers to *the monkey*, in sentence (2) to *the banana*, and in sentence (3), to the abstract notion of time. If we wish to translate these sentences into Spanish or German (languages which mark the gender of pronouns), anaphora resolution will be absolutely essential since, in these languages, pronouns take the gender of their antecedents. Therefore, in Spanish, we would obtain the following pronouns: (1) *éste* (in the masculine form since the antecedent—*the monkey*—is masculine), (2) *ésta* (feminine—*the banana*), and (3) an omitted pronoun (since the second clause of the sentence is impersonal in Spanish and does not need any subject). On the other hand, in German we would obtain: (1) *er* (masculine antecedent), (2) *sie* (feminine antecedent), and (3) *es* (neutral).

Besides these problems, originated by the gender of anaphoric expressions in different languages, there are other differences (that we have named discrepancies) which influence the process of translation of these expressions. These discrepancies have been previously studied by other authors. Mitkov & Schmidt (1998) present several problems to be taken into account in the translation of pronominal anaphors between different languages (German, French, English, Malay, and Korean); Nakaiwa & Ikeara (1992) treat the problem of the translation of elliptical constructions in a Japanese–English MT system; and Mitkov et al. (1994) and Geldbach (1999) present the discrepancies in an English–Korean MT system and a Russian–German MT system, respectively.

Another difference between languages is that of number discrepancies, in which certain nouns are referred to by a singular pronoun in one language and by a plural noun in the other. For example, the word *people* is plural in English, whereas in Spanish or German it is singular. Hence, in translations from English to Spanish, or from English to German, the plural pronoun will become a singular pronoun.

On the other hand, although in the majority of cases language-pairs pronouns in the source language are translated by target-language pronouns that correspond to the antecedent of the anaphor, there are some exceptions. In most of these cases, pronominal anaphors are simply omitted in the target language. For instance, in translations from English to Spanish, pronouns are frequently not translated because of the typical Spanish elliptical zero-subject construction. Other languages with typical zero constructions are Japanese, Italian, Thai, or Chinese.

In some languages, however, the pronoun is directly translated by its antecedent. For example, in English–Malay translations there is a tendency to replace the *it* pronoun with its antecedent, in which case the translator must first be able to identify the antecedent.

Some languages translate pronouns into different expressions, depending on the syntactic and semantic information of the antecedent. For example, in English–Korean translation pronouns can be elliptically omitted, or they can be translated into definite noun phrases, into their antecedent, or into different Korean pronouns.

All the above-mentioned problems in the translation of anaphoric expressions into a target language show that it is very important to carry out a detailed analysis of these expressions (including their resolution and the identification of the antecedent).

Because the majority of MT systems only handle one-sentence input, they usually cannot deal with anaphora resolution, and if they do, their successful operation usually does not go beyond the sentence level. In order to assess the deficiencies of MT systems, we analyzed the characteristics of different MT systems, with an emphasis on those characteristics related to anaphora resolution and translation into a target language. An overview of our analysis can be seen in Table 1.

| MT system | Strategy[a] | Restrict[b] | Partial[c] | Anaphor[d] | Corefer[e] | Zero[f] |
|---|---|---|---|---|---|---|
| Systran | Direct | No | Yes | Yes | No | Yes |
| Météo | Direct | Yes | No | No | No | No |
| SUSY | Transfer | No | No | Yes | No | No |
| Ariane | Transfer | No | No | Yes | No | Yes |
| Eurotra | Transfer | No | No | No | No | No |
| METAL | Transfer | No | Yes | Yes | No | Yes |
| Candide | Transfer | Yes | No | No | No | No |
| Inter-Nostrum | Transfer | Yes | Yes | No | No | No |
| IXA | Transfer | No | Yes | No | No | No |
| Episteme | Transfer | No | No | No | No | No |
| KANT | Interlingua | Yes | No | Yes | No | Yes |
| DLT | Interlingua | No | No | No | No | No |
| DLT (BKB) | Interlingua | No | No | Yes | No | No |
| Rosetta | Interlingua | No | No | No | No | No |
| CREST | Interlingua | Yes | No | Yes | Yes | Yes |
| $\mu$kosmos | Interlingua | Yes | No | No | No | No |

*a.* Strategy of translation: direct, transfer, or interlingua.
*b.* Restricted domain.
*c.* Partial parsing.
*d.* Resolution of intersentential anaphora.
*e.* Identification of co-reference chains.
*f.* Translation of zero pronouns into the target language.

Table 1: Characteristics of main MT systems

The table reflects a number of different system characteristics.

1. MT system. The MT systems studied included Systran (Toma, 1977; Wheeler, 1987); Météo (Chandioux, 1976, 1989); SUSY (Maas, 1977, 1987); Ariane (Boitet & Nédobejkine, 1981; Boitet, 1989); Eurotra (Varile & Lau, 1988; Allegranza, Krauwer, & Steiner, 1991); METAL (Bennet & Slocum, 1985; Thurmair, 1990); Candide (Berger et al., 1994); Inter-Nostrum (Canals-Marote et al., 2001a, 2001b); IXA (Díaz-Ilarraza, Mayor, & Sarasola, 2000, 2001); Episteme (Amores & Quesada, 1997; Quesada & Amores, 2000); KANT (Goodman, 1989; Nirenburg, 1989; Mitamura, Nyberg, & Carbonell, 1991); DLT (Witkam, 1983; Schubert, 1986); DLT with Bilingual Knowledge Bank (BKB) (Sadler, 1989); Rosetta (Appelo & Landsbergen, 1986; Landsbergen, 1987); CREST (Farwell & Helmreich, 2000); and $\mu$kosmos (Mahesh & Nirenburg, 1995a, 1995b).

2. Strategy of translation. This characteristic indicates the strategy used by the MT system in accordance with the existence of intermediate representations: direct, transfer, or interlingua.

3. Restricted domain. This characteristic tells if the texts of the source language are of a specific domain (restricted domain).

4. Partial parsing. This characteristic indicates if the MT system carries out a partial parsing of the source text by identifying only some constituents (noun phrases, prepositional phrases, etc.) and some relations between them.

5. Resolution of intersentential anaphora. This characteristic indicates whether the MT system resolves intersentential anaphora. If it does not, then the anaphoric expressions that have their antecedents in previous sentences will be incorrectly translated into the target language, in most of cases.

6. Identification of co-reference chains. This characteristic tells us if the co-reference chains of the source text are identified after resolving intersentential anaphora.

7. Translation of zero pronouns. This characteristic indicates if the MT system detects and resolves omitted pronouns (zero pronouns[1]) in the source language that are compulsory in the target language.

After analyzing the characteristics of the primary commercial MT systems, we concluded that there is no MT system that can work on unrestricted texts, resolve intersentential anaphora, identify the co-reference chains of the text, and translate zero pronouns into the target language after carrying out a partial parsing of the source text.

Unlike other systems, such as the KANT interlingua system, the Météo system, and the Candide system, among others, that are designed for well-defined domains, our interlingua MT approach, called AGIR (*Anaphora Generation with an Interlingua Representation*), works on unrestricted texts. Although we could have applied full parsing to these texts, we have instead utilized partial parsing, due to the unavoidable incompleteness of the grammar. This is a main difference between our system and other interlingua systems, such as the DLT system (which is based on a modification of Esperanto), the Rosetta system (which experiments with Montague semantics as the basis for an interlingua), the KANT system, and others.

After parsing and solving pronominal anaphora, an interlingua representation of the entire text is obtained. To do this, sentences are split into clauses, and a complex feature

1. This kind of pronouns will be presented in detail in Section 4.1.

structure based on semantic roles (agent, theme, etc.) is generated for each one. For each clause, the different semantic roles that appear will be identified and linked with one entity of the text. If there is an anaphor in the text, it will be linked with the entity that represents its antecedent. The AGIR's interlingua representation has been presented in more detail in (Peral, Palomar, & Ferrández, 1999; Peral & Ferrández, 2000a).

From the interlingua representation, the translation of the anaphor (including the intersentential anaphor), the detection of co-reference chains of the whole text, and the translation of Spanish zero pronouns into English have been carried out. AGIR has been designed to deal with all these issues which are hardly considered by most of the real MT systems. Furthermore, our approach can be used for other applications, for example, for cross-language information retrieval, summarization, etc.

It is important to note that although some of the above-mentioned MT systems resolve different problems, such as zero pronouns or pronominal anaphora, their results are not very satisfactory. Furthermore, we present some examples (extracted from the corpora used in the evaluation of our approach—see Section 4) of incorrect Spanish–English–Spanish translations of pronouns done by Systran[2] that AGIR does correctly[3]:

- (**S**) Siempre creí que a lo que yo aspiraba era a la comunicación perfecta con un hombre, o, mejor dicho, con el hombre, con ese príncipe azul de los sueños de infancia, un ser que sabría adivinarme hasta en los más menudos pliegues interiores. Ahora he aprendido no sólo que *[esa fusión]*$_i$ es imposible, sino además que *Ø*$_i$ es probablemente indeseable.

- (**E**) I always thought that to which I aspired I was to the perfect communication with a man, or, rather, with the man, that blue prince of the childhood dreams, a being who would know to guess to me until in slightest you fold interiors. Now I have learned not only that that fusion is impossible, but in addition that *he* is probably undesirable.

In this example, Systran incorrectly translates into English the zero pronoun of the last sentence of the paragraph, proposing the pronoun *he* instead of the pronoun *it* (the antecedent is the noun phrase *esa fusión—that fusion*). Our system proposed the correct pronoun. It is important to note that although the zero pronoun is identified by Systran, it is incorrectly solved and subsequently incorrectly translated.

- (**S**) Al pasar de *[la luminosidad]*$_i$ de la calle, y tal vez por contraste con *ella*$_i$, impresionaba la oscuridad interior y el vacío de la Catedral, en la que apenas había a la vista cuatro o cinco personas.

- (**E**) When happening of the luminosity of the street, and perhaps in contrast with *her*, it impressed the inner dark and the emptiness of the Cathedral, in which as soon as there were at sight four or five people.

2. A free trial of the commercial product SYSTRANLinks (copyright 2002 by SYSTRAN S.A.) has been used to translate between the English and Spanish languages all the corpora used in the evaluation of our approach. (URL = `http://w4.systranlinks.com/config`, visited on 06/22/2002).
3. In this paper, we have used the symbols (S) and (E) to represent Spanish and English texts, respectively. The symbol "*Ø*" indicates the presence of the omitted pronoun. In the examples, the pronoun and the antecedent have an index; co-indexing indicates co-reference between them.

In this case, Systran incorrectly translates into English the pronoun *ella* (with antecedent *la luminosidad—the luminosity*), by proposing the pronoun *her* instead of *it*.

- (**E**) If you have not already done so, unpack *[your printer]*$_i$ and the accessory kit that came with *it*$_i$.
- (**S**) Si usted no ha hecho ya así pues, desempaquete su impresora y el kit de accesorios eso vino con *él*.

This example shows an incorrect English–Spanish translation of the pronoun *it*. In this case, the pronoun is incorrectly solved (the antecedent is the noun phrase *your printer*, feminine) and then it is incorrectly translated (pronoun *él*—masculine—instead of pronoun *ésta*—feminine).

All the above examples illustrate that the translation of pronouns could be notably improved if their antecedents were correctly identified and, subsequently, pronouns were translated into the target language.

## 3. AGIR's Analysis Module

AGIR system architecture is based on the general architecture of an MT system that uses an interlingua strategy. Translation is carried out in two stages: (1) from the source language to the interlingua, and (2) from the interlingua into the target language. Modules for analysis are independent from modules for generation. Although our present work has only studied the Spanish and English languages, our approach can be easily extended to other languages, for exampe, to multilingual system, in the sense that any analysis module can be linked to any generation module.

In AGIR the analysis is carried out using SUPAR (*slot unification parser for anaphora resolution*) (Ferrández et al., 1999). SUPAR is a computational system that focuses on anaphora resolution. It can deal with several kinds of anaphora, such as pronominal anaphora, one-anaphora, or definite descriptions[4]. The SUPAR's input is a grammar defined by means of the grammatical formalism SUG (*slot unification grammar*). A translator that transforms SUG rules into Prolog clauses has been developed. This translator provides a Prolog program that will parse each sentence. SUPAR can perform either a full or a partial parsing of the text with the same parser and grammar. In this study, partial-parsing techniques have been utilized due to the unavoidable incompleteness of the grammar and the use of unrestricted texts (corpora) as input.

The analysis of the source text is carried out in several steps. The first step of the analysis module is the lexical and morphological analysis of the input text. Because of the use of unrestricted texts as input, the system obtains the lexical and morphological information of the texts' lexical units from the output of a part-of-speech (POS) tagger. The word as it appears in the corpus, its lemma, and its POS tag (with morphological information) is supplied for each lexical unit in the corpus.

4. *One-anaphora* has the following structure in English: a determiner and the pronoun *one* with some premodifiers or postmodifiers (*the red one*; *the one with the blue bow*). This kind of anaphors in Spanish consists of noun phrases in which the noun has been omitted (*el rojo*; *el que tiene el lazo azul*). In *definite descriptions*, anaphors are formed by definite noun phrases that refer to objects that are usually uniquely determined in the context.

The next step is the parsing of the text (which includes the lexical and morphological information extracted in the previous stage). Before applying the parsing, the text is split into sentences. The output will be the slot structure (SS) that stores the necessary information[5] for the subsequent stages.

In the third step, a module of word-sense disambiguation (WSD) is used to obtain a single sense for the different texts' lexical units. The lexical resources, WordNet (Miller, Beckwith, Fellbaum, Gross, & Miller, 1990) and EuroWordNet (Vossen, 1998), have been used in this stage[6].

The SS, enriched with the information from the previous steps, will be the input for the next step, in which NLP problems (anaphora, extraposition, ellipsis, etc.) will be treated and solved. In this work, we have focused on the resolution of NLP problems related to pronominal anaphora. After this step, a new slot structure (SS') is obtained. In this new structure, the correct antecedent—chosen from the possible candidates after applying a method based on constraints and preferences (Ferrández et al., 1999)—for each anaphoric expression will be stored along with its morphological and semantic information. The new structure SS' will be the input for the final step of the analysis module.

In the last step, AGIR generates the interlingua representation of the entire text. This is the main difference between AGIR and other MT systems, which process the input text sentence by sentence. The interlingua representation will allow the correct translation of the intersentential and intrasentential pronominal anaphora into the target language. Moreover, AGIR allows the identification of co-reference chains of the text and their subsequent translation into the target language.

The interlingua representation of the input text is based on the clause as the main unit of this representation. Once the text has been split into clauses, AGIR uses a complex feature structure for each clause. This structure is composed of semantic roles and features extracted from the SS of the clause. The notation we have used is based on the one used in KANT interlingua.

It is important to emphasize that the interlingua lexical unit has been represented in AGIR using the word and its correct sense in WordNet. After accessing the ILI (interlingual-index) module of EuroWordNet, we will be able to generate the lexical unit into the target language.

Once the semantic roles have been identified, the interlingua representation will store the clauses with their features, the different entities that have appeared in the text and the relations between them (such as anaphoric relations). This representation will be the input for the generation module.

5. The SS stores the following information for each constituent: constituent name (NP, PP, etc.), semantic and morphological information, discourse marker (identifier of the entity or discourse object), and the SS of its subconstituents.
6. In the evaluation of our approach, we have only used an English corpus (SemCor) where all content words are annotated with their WordNet sense; this sense has been used to identify the semantic category of the word. The remaining corpora do not have information about the senses of the content words; therefore, a set of heuristics has been used to identify their semantic categories. Currently, a WSD module (Montoyo & Palomar, 2000) is being developed in our Research Group, which will be incorporated into our system in the future.

## 4. Resolution of NLP Problems in AGIR

The fourth stage of the AGIR's analysis module allows for the resolution of NLP problems. Our present work focuses on the resolution of NLP problems related to pronominal anaphora in the source language so as to translate these anaphoric expressions correctly into the target language. We are only describing the translation of anaphoric, third-person, personal pronouns into the target language. Therefore, we have only focused on the discrepancies between Spanish and English in the treatment of these pronouns. In the next two subsections, we will describe the syntactic discrepancies treated and solved in AGIR (Spanish zero pronouns) and the anaphora resolution module of the system.

### 4.1 Elliptical Zero-Subject Constructions (Zero Pronouns)

The Spanish language allows for the omission of the pronominal subject of the sentences. These omitted pronouns are usually called zero pronouns. Whereas in other languages (e.g., in Japanese), zero pronouns may appear in either the subject's or the object's grammatical position, in Spanish texts zero pronouns only appear in the position of the subject.

In MT systems, the correct detection and resolution of zero pronouns in the source language is of crucial importance if these pronouns are compulsory in the target language. In the following example, a Spanish sentence that contains a zero pronoun and its translation into English with the equivalent compulsory pronoun are shown.

- (**S**) *[Ese hombre]*$_i$ era un boxeador profesional. $\O_i$ Perdió únicamente dos combates.
- (**E**) *[That man]*$_i$ was a professional boxer. *He*$_i$ only lost two fights.

We should remark that zero pronouns can also occur in English, although they appear less frequently, since they usually are used in coordinated sentences in which the zero pronoun usually refers to the subject of the sentence. Although zero pronouns have already been studied in other languages, such as Japanese—with a resolution percentage of 78% in the work of (Okumura & Tamura, 1996), they have not yet been studied in Spanish texts. (Ferrández & Peral, 2000) has presented the first algorithm for Spanish zero-pronoun resolution. Basically, in order to translate Spanish zero pronouns into English, they must first be located in the text (ellipsis detection) and then resolved (anaphora resolution) (Peral & Ferrández, 2000b):

- Zero-pronoun detection. In order to detect zero pronouns, sentences should be divided into clauses, since the subject can only appear between the clause constituents. After that, a noun-phrase (NP) or a pronoun is sought, for each clause, through the clause constituents on the lefthand side of the verb, unless it is imperative or impersonal. Such an NP or pronoun must agree in person and number with the verb of the clause.

- Zero-pronoun resolution. After the zero pronoun has been detected, our computational system inserts the pronoun in the position in which it has been omitted. This pronoun will be detected and resolved in the following module of anaphora resolution. Person and number information is obtained from the clause verb. Sometimes, in Spanish, the gender information of the pronoun can be obtained from the object when the verb is copulative. In these cases, the subject must agree in gender and number with its object whenever the object can have either a masculine or feminine linguistic form.

#### 4.1.1 Evaluation

To evaluate this task, two experiments were performed: an evaluation of zero-pronoun detection and an evaluation of zero-pronoun resolution. In both experiments the method was tested on two kinds of corpora. In the first instance, we used a portion of the LEXESP[7] corpus that contains a set of thirty-one documents (38,999 words) from different genres and written by different authors. The LEXESP corpus contains texts of different styles and on different topics (newspaper articles about politics, sports, etc.; narratives about specific topics; novel fragments; etc.). In the second instance, the method was tested on a fragment of the Spanish version of Blue Book (BB) corpus (15,571 words), a technical manual that contains the handbook of the International Telecommunications Union (CCITT) published in English, French, and Spanish. Both corpora are automatically tagged by different taggers.

We randomly selected a subset of the LEXESP corpus (three documents —6,457 words) and a fragment of the Blue Book corpus (4,723 words) as training corpora. The remaining fragments of the corpora were reserved for test data.

It is important to emphasize that all the tasks presented in this paper were automatically evaluated after the annotation of each pronoun (including zero pronouns). To do so, each anaphoric, third-person, personal pronoun was annotated with the information about its antecedent and its translation into the target language. Furthermore, co-reference chains were identified. The annotation phase was accomplished in the following manner: (1) two annotators (native speakers) were selected for each language, (2) an agreement was reached between the annotators with regard to the annotation scheme, (3) each annotator annotated the corpora, and (4) a reliability test (Carletta et al., 1997) was done on the annotation in order to guarantee the results. The reliability test used the *kappa* statistic that measures agreement between the annotations of two annotators in making judgments about categories. In this way, the annotation is considered a classification task consisting of defining an adequate solution among the candidate list. According to Carletta et al. (1997), a $k$ measurement such as $0.68 < k < 0.80$ allows us to draw encouraging conclusions, and a measurement $k > 0.80$ means there is total reliability between the results of the two annotators. In our tests, we obtained a *kappa* measurement of 0.83. Therefore, we consider the annotation obtained for the evaluation to be totally reliable.

#### 4.1.2 Evaluation of Zero-Pronoun Detection

In the evaluation of zero-pronoun detection, the training phase was used to carry out modifications in the grammar in order to improve the processes of partial parsing and clause splitting. After this training, we conducted a blind test over the entire test corpus. To achieve this sort of evaluation, several different subtasks may be considered. First, each verb must be detected. This task is easily accomplished since both corpora have been pre-

7. The LEXESP corpus belongs to the project of the same name, carried out by the Psychology Department of the University of Oviedo and developed by the Computational Linguistics Group of the University of Barcelona, with the collaboration of the Language Processing Group of the Catalonia University of Technology, Spain.

viously tagged. The second task is to classify the verbs into two categories: (a) verbs whose subjects have been omitted, and (b) verbs whose subjects have not. The obtained results with the LEXESP and Blue Book corpora appear in Table 2.

| | Verbs with subject omitted | | | | | | Verbs with subject *not* omitted | | | | | |
|---|---|---|---|---|---|---|---|---|---|---|---|---|
| | $1^{st}$ | P(%) | $2^{nd}$ | P(%) | $3^{rd}$ | P(%) | $1^{st}$ | P(%) | $2^{nd}$ | P(%) | $3^{rd}$ | P(%) |
| LX | 240 | 96.7 | 54 | 98.1 | 1,227 | 97.1 | 31 | 71 | 17 | 94.1 | 1,085 | 83.3 |
| | PRECISION = 97.1% | | | | | | PRECISION = 83.1% | | | | | |
| BB | 0 | 0 | 0 | 0 | 121 | 97.5 | 0 | 0 | 0 | 0 | 351 | 82 |
| | PRECISION = 97.5% | | | | | | PRECISION = 82.0% | | | | | |
| **GLOBAL PRECISION = 90.4%** | | | | | | | | | | | | |

Table 2: Zero-pronoun detection, evaluation phase

The table is divided into two parts, corresponding to categories (a) and (b) previously mentioned. For each category, the number of verbs in first, second, and third person, together with their precision (**P**), are represented. Precision was defined as the number of verbs correctly classified (subject omitted or not) divided by the total number of verb classifications attempted for each type. For example, in the LEXESP corpus 1,227 verbs in the third person with their subjects omitted were classified, and the precision obtained was 97.1%.

**Discussion.** In the detection of zero pronouns the following results were obtained: for the LEXESP corpus, precisions of 97.1% and 83.1% were obtained for verbs whose subjects were omitted or were not omitted, respectively; for the BB corpus, precisions of 97.5% and 82% were obtained. For both corpora, an overall precision of 90.4% (2,825 out of a total of 3,126) was obtained for this task.

From these results, we have extracted the following conclusions:

- There are no meaningful differences between the results obtained with each corpus.
- The BB corpus has no verbs in either the first or second person. This is explained by considering the style of the corpus: it is a technical manual which usually consists of a series of isolated definitions done by the writer.
- The rate of precision for the detection of verbs whose subjects are not omitted is lower (approximately 15%) than for the detection of verbs whose subjects are omitted. There are several reasons for this:
  - The POS tagger does not identify impersonal verbs. This problem has been partly resolved heuristically, by the choice of impersonal verbs (e.g., *llover—to rain*), but it cannot be resolved for all impersonal verbs. For example, the verb *ser* (*to be*) is not usually impersonal, but it can be in certain constructions (e.g., *Es hora de desayunar—It is time to have breakfast*).
  - The ambiguity and the unavoidable incompleteness of the grammar affects the process of clause splitting, and therefore affects the detection of the possible subject for the clause on the lefthand side of the verb.

Since ours is the first study done specifically on Spanish texts and since the design of the detection stage mainly depends upon the structure of the language in question, we have not compared our results with those of other published works. Such comparisons would prove to be insignificant[8].

Finally, it is important to emphasize the importance of this phenomenon in Spanish. Specifically, in both corpora, the subject is omitted in 52.5% (1,642 out of 3,126) of the verbs. Furthermore, this phenomenon is even more important in narrative texts (57.3% in the LEXESP corpus) than in the technical manuals (25.6% in the BB corpus). These percentages show the importance of correctly detecting these kinds of pronouns in an MT system so as to conveniently translate them into the target language.

#### 4.1.3 Evaluation of Zero-Pronoun Resolution

After zero pronouns have been detected, they are then resolved in the subsequent module of anaphora resolution (explained in the following subsection). Basically, an algorithm that combines different kinds of knowledge by distinguishing between constraints and preferences is used (Ferrández et al., 1999; Palomar et al., 2001).

The set of constraints and preferences presents two basic differences between zero-pronoun and pronominal anaphora resolution:

1. Zero-pronoun resolution has the constraint of agreement only in person and number, whereas pronominal anaphora resolution also requires gender agreement.
2. Two new preferences to solve zero pronouns are used: (a) preference is given to candidates in the same sentence as the anaphor that have also been the solution of a zero pronoun in the same sentence as the anaphor, and (b) in the case where the zero pronoun has gender information, preference is given to those candidates that agree in gender.

In evaluating zero-pronoun resolution so as to obtain the best order of preferences (one that produces the best performance), we used the training phase to identify the importance of each kind of knowledge. To do this, we analyzed the antecedent for each pronoun in the training corpora, and we identified their configurational characteristics with reference to the pronoun (e.g., *if the antecedent was a proper noun*, *if the antecedent was an indefinite NP*, *if the antecedent occupied the same position with reference to the verb as the anaphor —before or after*, etc.). Subsequently, we constructed a table that showed how often each configurational characteristic was valid for the solution of a particular pronoun (e.g., the solution of a zero pronoun was a proper noun 63% of the time, for a reflexive pronoun, it was a proper noun 53% of the time, etc.). In this way, we were able to define the different patterns of Spanish pronoun resolution and apply them in order to obtain the evaluation results that are presented in this paper. The order of importance was determined by first sorting the preferences according to the percentage of each configurational characteristic;

8. In order to compare our system with other systems, in Section 6.2 we evaluate pronoun translation (including zero pronouns) between Spanish and English using the commercial product SYSTRANLinks and the Spanish LEXESP corpus. The evaluation highlights the deficiencies of zero-pronoun detection, resolution, and translation (out of 559 anaphoric, third-person, zero pronouns in the LEXESP corpus, only 266 were correctly translated into English—a precision of only 47.6%).

that is, preferences with higher percentages were considered more important than those with lower percentages. After several experiments on the training corpora, an optimal order for each type of anaphora was obtained. Since in this phase we processed texts from different genres and by different authors, we can state that the final set of preferences obtained and their order of application can be used with confidence on any Spanish text.

After the training, we conducted a blind test over the entire test corpus, the results for which are shown in Table 3.

| | Cataphoric | Exophoric | Anaphoric | | |
|---|---|---|---|---|---|
| | | | Correct | Total | P(%) |
| LEXESP | 640 | 28 | 455 | 559 | 81.4 |
| BB | 76 | 8 | 30 | 37 | 81.1 |
| **TOTAL** | **716** | **36** | **485** | **596** | **81.4** |

Table 3: Zero-pronoun resolution, evaluation phase

It is important to mention here that out of 3,126 verbs in these corpora, 1,348 (Table 2) are zero pronouns in the third person and will be resolved. In Table 3 we present a classification of these third-person zero pronouns, which has been conveniently divided into three categories:

1. Cataphoric. This category is comprised of those zero pronouns whose antecedents, that is, the clause subjects, come after the verb. For instance, in the following Spanish sentence $Ø_i$ *Compró* [un niño]$_i$ *en el supermercado* ([A boy]$_i$ *bought in the supermarket*), the subject, *un niño* (*a boy*), appears after the verb, *compró* (*bought*). These kinds of verbs are quite common in Spanish (**P** = 53.1%, 716 out of 1,348), as can be seen in Table 3, and represents one of the main difficulties in resolving anaphora in Spanish: the structure of a sentence is more flexible than in English. These represent intonationally marked sentences, where the subject does not occupy its usual position in the sentence, that is, before the verb. Cataphoric zero pronouns will not be resolved in AGIR, since semantic information is needed to be able to discard all of their antecedents and *to give preference to* those that appear within the same sentence and clause after the verb.

   For example, the sentence Ø *Compró un regalo en el supermercado* ([He]$_{Ø}$ *bought a present in the supermarket*) has the same syntactic structure as the previous sentence, i.e., verb, NP, and PP, where the object function of the NP can only be distinguished from the subject by means of semantic knowledge.

2. Exophoric. This category consists of those zero pronouns whose antecedents do not appear linguistically in the text (they refer to items in the external world rather than things referred to in the text). Exophoric zero pronouns will not be resolved by the system.

3. Anaphoric. This category is comprised of those zero pronouns whose antecedents are found before the verb. These kinds of pronouns will be resolved by our system.

In Table 3 the numbers of cataphoric, exophoric, and anaphoric zero pronouns for each corpus are shown. For anaphoric pronouns, the number of pronouns correctly solved as well as the obtained precision, **P** (number of pronouns correctly solved divided by the number of solved pronouns) is presented. For example, in the LEXESP corpus, there are 640 cataphoric, 28 exophoric, and 559 anaphoric zero pronouns. From these anaphoric pronouns, only 455 were correctly solved, giving a precision of 81.4%.

**Discussion.** In zero-pronoun resolution, the following results have been obtained: LEXESP corpus, **P** = 81.4%; BB corpus, **P** = 81.1%. For the combined corpora, an overall precision for this task of 81.4% (485 out of 596) was obtained. The overall recall, **R** (the number of pronouns correctly solved divided by the number of real pronouns) obtained was 79.1% (485 out of 613).

From these results, we have extracted the following conclusions:

- There are no meaningful differences between the results obtained from each corpus.
- Errors in the zero-pronoun-resolution stage are originated by different causes:
  - exceptions in the application of preferences that imply the selection of an incorrect antecedent as solution of the zero pronoun (64% of the global mistakes)
  - the lack of semantic information[9], causing an error rate of 32.4%
  - mistakes in the POS tagging (3.6%)

Since the results provided by other works have been obtained for different languages (English), texts, and sorts of knowledge (e.g., Hobbs and Lappin full parse the text), direct comparisons are not possible. Therefore, in order to accomplish this comparison, we have implemented some of these approaches in SUPAR[10], adapting them for partial parsing and Spanish texts. Although these approaches were not proposed for zero pronouns and the comparison will not be fully fair, we have implemented them since that is the only way to compare our proposal directly with some well-known anaphora-resolution algorithms.

We have also compared our system with the typical baseline of proximity preference (i.e., the antecedent that appears closest to the anaphora is chosen from among those that satisfy the constraints—morphological agreement and syntactic conditions). We have also compared our system with the baseline presented by Hobbs (1978)[11] and Lappin & Leass' method (Lappin & Leass, 1994). Moreover, we also compared our proposal with centering approach by implementing functional centering (Strube & Hahn, 1999). The precisions obtained with these different approaches and AGIR are shown in Table 4. As can be seen, the precision obtained in AGIR is better than those obtained using the other proposals.

9. It is important to mention here that semantic information was not available for the Spanish corpora.
10. A detailed study of these implementations in SUPAR is presented in Palomar et al. (2001).
11. Hobbs's baseline is frequently used to compare most of the work accomplished on anaphora resolution. Hobbs's algorithm does not work as well as ours because it carries out a full parsing of the text. Furthermore, the manner in which the syntactic tree is explored using Hobbs's algorithm is not the best one for Spanish, since it is nearly a free-word-order language.

| | Proximity | Hobbs | Lappin | Strube | AGIR |
|---|---|---|---|---|---|
| LEXESP | 54.9 | 60.4 | 66.0 | 59.7 | **81.4** |
| BB | 48.6 | 62.2 | 67.6 | 59.5 | **81.1** |

Table 4: Zero-pronoun resolution in Spanish, comparison of AGIR with other approaches

### 4.2 The Anaphora-Resolution Module

The anaphora-resolution module used in AGIR is based on the module presented in (Ferrández et al., 1999; Palomar et al., 2001) for the SUPAR system. The algorithm identifies noun phrase (NP) antecedents of personal, demonstrative, reflexive, and zero pronouns in Spanish. It identifies both intrasentential and intersentential antecedents and is applied to the syntactic analysis generated by SUPAR. It also combines different forms of knowledge by distinguishing between constraints and preferences. Whereas constraints are used as combinations of several kinds of knowledge (lexical, morphological, and syntactic), preferences are defined as a combination of heuristic rules extracted from a study of different corpora.

A constraint defines a property that must be satisfied in order for any candidate to be considered as a possible solution of the anaphor. The constraints used in the algorithm are the following: morphological agreement (person, gender, and number) and syntactic conditions on NP-pronoun non-co-reference.

A preference is a characteristic that is not always satisfied by the solution of an anaphor. The application of preferences usually involves the use of heuristic rules in order to obtain a ranked list of candidates. Some examples of preferences used in our system are the following: (a) antecedents that are in the same sentence as the anaphor, (b) antecedents that have been repeated more than once in the text, (c) antecedents that appear before their verbs (i.e., the verb of the clause in which the antecedent appears), (d) antecedents that are proper nouns, (e) antecedents that are an indefinite NP, and so on.

In order to solve pronominal anaphors, they must be first located in the text (anaphora detection) and then resolved (anaphora resolution):

- Anaphora detection. In the algorithm, all the types of anaphors are identified from left to right as they appear in the sentence's slot structure obtained after the partial parsing. To identify each type of pronoun, the information stored in the POS tags has been used. In the particular case of zero pronouns, they have been detected in a previous stage, as previously described.

- Anaphora resolution. After the anaphor has been detected, the corresponding method, based on constraints and preferences, is applied to solve it. Each type of anaphor has its own set of constraints and preferences, although they all follow the same general algorithm: constraints are applied first, followed by preferences. Constraints discard some of the candidates, whereas preferences simply sort the remaining candidates.

#### 4.2.1 Evaluation

In evaluating the algorithm for anaphora resolution[12], we looked at pronominal anaphora resolution in Spanish and English, respectively. For the Spanish evaluation, the method was tested on the portion of the LEXESP corpus previously used to evaluate zero-pronoun detection and resolution. For English, we tested the method on two kinds of corpora. In the first instance, we used a portion of the SemCor collection—presented in (Landes, Leacock, & Tengi, 1998)—which contains a set of eleven documents (23,788 words) in which all content words are annotated with the most appropriate WordNet sense. The SemCor corpus contains texts about different topics (law, sports, religion, nature, etc.) and was written by different authors. In the second instance, the method was tested on a portion of the MTI[13] corpus, which contains seven documents (101,843 words). The MTI corpus contains computer science manuals on different topics (commercial applications, word processing applications, device instructions, etc.). Both English corpora are automatically tagged by different taggers.

We randomly selected a subset of the SemCor corpus (three documents—6,473 words) and another subset of the MTI corpus (two documents—24,264 words) as training corpus. The remaining fragments of the corpora were reserved for test data.

In the two tasks, the training phase was used to identify the importance of each kind of knowledge to obtain the optimal order of the preferences.

#### 4.2.2 Evaluation of Anaphora Resolution in Spanish

An evaluation of the algorithm for anaphora resolution in Spanish has been given in detail in the work of Palomar et al. (2001). In this paper, we present the obtained results of the evaluation of this task in AGIR over a different portion of the LEXESP corpus. Furthermore, non-anaphoric complement pronouns, that is, complement pronouns that appear next to the previous indirect object when it has been moved from its theoretical place after the verb (*A Pedro*$_i$ le$_i$ *vi ayer—I saw Pedro yesterday*), were not resolved because this kind of pronoun does not appear in the English translation. For these reasons, the results of the two works are slightly different.

After the training phase, the algorithm was evaluated over the test corpus. In this evaluation, only lexical, morphological, and syntactic information was used. Table 5 shows the results of this evaluation.

| | Comp | P(%) | Ref | P(%) | PP notPP | P(%) | PP inPP | P(%) | Total | P(%) Total |
|---|---|---|---|---|---|---|---|---|---|---|
| LEXESP | 98 | 82.6 | 105 | 92.4 | 71 | 70.4 | 46 | 76.1 | 320 | 82.2 |

Table 5: Anaphora resolution in Spanish, evaluation phase

12. As previously mentioned, only anaphoric, third-person, personal pronouns will be resolved in order to translate them into the target language.
13. This corpus was provided by the Computational Linguistics Research Group of the School of Humanities, Languages and Social Studies, University of Wolverhampton, England. The corpus is anaphorically annotated indicating the anaphors and their correct antecedents.

In Table 5 the occurrences of personal pronouns in the LEXESP corpus are shown. The different types are: *Comp* (complement personal pronouns), *Ref* (reflexive pronouns), *PPnotPP* (personal pronouns not included in a prepositional phrase), and *PPinPP* (personal pronouns included in a prepositional phrase). For each type, the obtained precision, **P** (the number of pronouns correctly solved divided by the number of solved pronouns), is shown. The last two columns represent the total number of personal pronouns and the obtained precision.

**Discussion.** In pronominal anaphora resolution in Spanish, we obtained a precision of 82.2% (263 out of 320). The recall, **R** (number of pronouns correctly solved divided by the number of real pronouns), obtained was of 79% (263 out of 333).

After analyzing the results, the following conclusions were extracted:

- In the resolution of reflexive pronouns, a high precision (92.4%) was obtained. This higher percentage is because the antecedent of these pronouns is usually the closest NP to the pronoun and it is in the same sentence. Therefore, after applying preferences, few errors are produced.

- Analyzing the errors in the remaining pronouns, it is important to mention the complexity of the LEXESP corpus itself. It consists of several narrative documents, sometimes with a very complex style, with long sentences (with an average of 24.6 words per sentence). This implies a large number of candidates per anaphor after applying constraints (an average of 16.6).

- Errors were originated by different causes:

  - exceptions in the application of preferences (66.7% of the global mistakes)
  - a lack of semantic information (29.8%)
  - mistakes in the POS tagging (3.5%)

We compared our proposal with the approaches previously presented in the evaluation of zero-pronoun resolution. As shown in Table 6, the precision obtained using AGIR is better than those for the other proposals.

| | Proximity | Hobbs | Lappin | Strube | AGIR |
|---|---|---|---|---|---|
| LEXESP | 52.5 | 65.3 | 73.3 | 68.3 | **82.2** |

Table 6: Anaphora resolution in Spanish, comparison of AGIR with other approaches

### 4.2.3 Evaluation of Anaphora Resolution in English

The algorithm for anaphora resolution in English is based on the one developed for Spanish, and it has been conveniently adapted for English. The main difference between the two algorithms consists in a different order of the preferences obtained after the training phase. After this phase, we extracted the following conclusions:

- Spanish has more morphological information than English. As a consequence, morphological constraints in Spanish discard more candidates than constraints in English.
- Spanish is a nearly free-order language, in which the different constituents of a sentence (subject, object, etc.) can appear almost at any position. For this reason, the preference of *syntactic parallelism* has a more important role in the anaphora-resolution method in English than in Spanish.
- Spanish sentences are usually longer than English ones. This fact implies more candidates for Spanish anaphors than for English ones.

After the training phase, the algorithm was evaluated over the test corpus. In the evaluation phase, two experiments were carried out. In the first experiment, only lexical, morphological, and syntactic information was used. The obtained results with the SemCor and MTI corpora appear in Table 7.

| | He | She | It | They | Him | Her | Them | Corr | Total | P(%) |
|---|---|---|---|---|---|---|---|---|---|---|
| SEMCOR | 116 | 10 | 38 | 50 | 34 | 0 | 6 | 175 | 254 | 68.9 |
| MTI | 1 | 0 | 347 | 56 | 0 | 0 | 66 | 361 | 470 | 76.8 |

Table 7: Anaphora resolution in English, evaluation phase: experiment 1

The table shows the number of pronouns (classified by type) for the different corpora. The last three columns represent the number of correctly solved pronouns, the total number of pronouns, and the obtained precision, respectively. For instance, in the MTI corpus a precision of 76.8% was obtained.

**Discussion.** In pronominal anaphora resolution in English, the following results were obtained in the first experiment: SemCor corpus, $\mathbf{P}$ = 68.9%, $\mathbf{R}$ = 66%; MTI corpus, $\mathbf{P}$ = 76.8%, $\mathbf{R}$ = 72.9%.

From these results, we have extracted the following conclusions:

- The types of pronouns vary considerably according to the corpus. In the SemCor corpus, 15% of the pronouns are occurrences of the *it* pronoun, whereas in the MTI corpus this percentage is 73.8%. This fact is explained by the kind and domain of each corpus. The SemCor is a corpus with a narrative style which contains a lot of *person* entities[14] that are referred to in the text with the use of personal pronouns (*he*, *she*, and *they*). On the other hand, the MTI corpus is a collection of technical manuals that contains almost no *person* entities. Rather, most references are to *object* entities, using *it* pronouns.
- In the SemCor corpus, errors originated from different causes:
  - The lack of semantic information caused 57% of the global mistakes. There were seventeen mistakes in the resolution of *it* pronouns, in which the system proposed

14. If we use a basic ontology based on semantic features, at the top level, entities could be classified into three main categories: *person*, *animal*, and *object*.

a *person* entity as solutions for these pronouns. On the other hand, twenty-eight occurrences of the pronouns *he*, *she*, *him*, and *her* were incorrectly solved due to the system proposing an *object* or *animal* entity as the solution.

- There were exceptions in the applications of preferences (38%), mainly due to the existence of a large number of candidates compatible with the anaphor[15].
- There were mistakes in the POS tagging (5%).

- In the MTI corpus, errors were mainly produced in the resolution of *it* pronouns (73.4% of the global mistakes). The *it* pronoun lacks gender information (it is valid for masculine and feminine) and subsequently there are a lot of candidates per anaphor[16]. This fact originates errors in the application of preferences. The remaining errors are originated by the lack of semantic information.

After analyzing the results, it was observed that the precision of the SemCor corpus was approximately 8% lower than that for the MTI corpus. The errors in the SemCor corpus mainly originated with the lack of semantic information. Therefore, in order to improve the obtained results, a second experiment was carried out with the addition of semantic information.

The modifications to the second experiment were the following:

- Two new semantic constraints—presented in (Saiz-Noeda, Peral, & Suárez, 2000)—were added to the morphological and syntactic constraints:

    1. The pronouns *he*, *she*, *him*, and *her* must have as the antecedents *person* entities.
    2. The pronoun *it* must have as its antecedent a *non-person* entity.

  To apply these new constraints, the twenty-five *top concepts* of WordNet (the concepts at the top level in the ontology) were grouped into three categories: *person*, *animal*, and *object*. Subsequently, WordNet was consulted with the head of each candidate, and thus the semantic category of the antecedent was obtained.

- This experiment was exclusively carried out with the SemCor corpus because it is the only one in which content words are annotated with their WordNet sense.

Table 8 shows the number of pronouns (classified by type) for the different corpora after these changes were incorporated.

As shown in Table 8, the addition of the two simple semantic constraints resulted in considerable improvement in the obtained precision (approximately 18%) for the SemCor corpus. We concluded that the use of semantic information (such as new constraints and preferences) in the process of anaphora resolution will improve the results obtained.

15. The sentences of the SemCor corpus are very long (with an average of 24.3 words per sentence). This fact implies a large number of candidates per anaphor (an average of 15.2) after applying constraints.
16. The sentences of the MTI corpus are not very long (with an average of 15.5 words per sentence). However, the candidates per anaphor, after applying constraints, are high (an average of 13.6).

| | He | She | It | They | Him | Her | Them | Corr | Total | P(%) |
|---|---|---|---|---|---|---|---|---|---|---|
| SEMCOR | 116 | 10 | 38 | 50 | 34 | 0 | 6 | 220 | 254 | 86.6 |
| MTI | 1 | 0 | 347 | 56 | 0 | 0 | 66 | 361 | 470 | 76.8 |

Table 8: Anaphora resolution in English, evaluation phase: experiment 2

Finally, Table 9 compares anaphora resolution using AGIR with the other approaches previously presented[17]. It is important to emphasize the high percentages obtained using our system and Hobbs's method in the SemCor corpus; both systems incorporate semantic information[18] into their methods using semantic constraints (selectional restrictions), whereas none of the other authors incorporate semantics in their approaches.

| | Proximity | Hobbs | Lappin | Strube | AGIR |
|---|---|---|---|---|---|
| SEMCOR | 37.0 | 81.9 | 59.4 | 59.4 | **86.6** |
| MTI | 54.9 | 66.0 | 75.1 | 63.2 | **76.8** |

Table 9: Anaphora resolution in English, comparison of AGIR with other approaches

## 5. AGIR's Generation Module

The interlingua representation of the source text is taken as input for the generation module. The output of this module is the target text, that is, the representation of the source text's meaning with words of the target language.

The generation phase is split into two modules: syntactic generation and morphological generation. Although the approach presented here is multilingual, we have focused on the generation into the Spanish and English languages.

### 5.1 Syntactic Generation

In syntactic generation, the interlingua representation is converted by *transformational rules* into an ordered surface-structure tree, with appropriate labeling of the leaves with target language grammatical functions and features. The basic task of syntactic generation is to order constituents in the correct sequence for the target language. However, the aim of this work is only the translation of pronominal anaphora into the target language, so we have only focused on the discrepancies between the Spanish and English languages in the translation of the pronoun.

In syntactic generation, Spanish elliptical zero-subject constructions were studied. This phenomenon was conveniently treated and solved in the analysis module. Therefore, all the

17. As mentioned earlier, all the results presented here were automatically obtained after the anaphoric annotation of each pronoun. After the tagging and the partial parsing of the source text, pronominal anaphora were resolved and translated into the target language. None of the intermediate outputs needed to be adjusted manually in order to be processed subsequently.
18. Hobbs proposed the use of semantic information using selectional restrictions as a straightforward extension of his method in order to improve the obtained results in anaphora resolution.

necessary information to translate these constructions has been stored in the interlingua representation.

### 5.2 Morphological Generation

The final stage of the generation module is morphological generation, in which we mainly have to treat and solve number and gender discrepancies in the translation of pronouns.

#### 5.2.1 Number Discrepancies

This problem is generated by the discrepancy between words of different languages that express the same concept. These words can be referred to a singular pronoun in the source language and to a plural pronoun in the target language. In order to take into account number discrepancies in the translation of the pronoun into English or Spanish, a set of morphological (number) rules is constructed. The lefthand side of the number rule contains the interlingua representation of the pronoun, whereas the righthand side contains the pronoun in the target language.

#### 5.2.2 Gender Discrepancies

Gender discrepancies come from existing morphological differences between different languages. For instance, English has less morphological information than Spanish. The English plural personal pronoun *they* can be translated into the Spanish pronouns *ellos* (masculine) or *ellas* (feminine); the singular personal pronoun *it* can be translated into *él/éste* (masculine) or *ella/ésta* (feminine), etc. In order to take into account such gender discrepancies in the translation of the pronoun into English or Spanish, a set of morphological (gender) rules was constructed.

## 6. Evaluation of the Generation Module

In this step, we tested the AGIR's generation module by evaluating the translation of English pronouns into Spanish, and the translation of Spanish pronouns into English.

As mentioned earlier, the generation module takes the interlingua representation as input. Previously, Spanish zero pronouns were detected (90.4% **P**) and resolved (81.4% **P**), and anaphoric third-person personal pronouns were resolved in Spanish (82.2% **P**) and English (86.6% **P** in the SemCor corpus with semantic information, and 76.8% **P** in the MTI corpus without semantic information). Non-referential uses of *it* pronouns were automatically detected, obtaining an 88.7% **P** on unrestricted texts[19].

19. In order to detect pleonastic *it* pronouns in AGIR, a set of rules, based on pattern recognition, that allow for the identification of this type of pronoun is constructed. These rules were based on the work of (Lappin & Leass, 1994; Paice & Husk, 1987; Denber, 1998), which dealt with this problem in a similar way. We have used the information provided by the POS tagger in order to improve the detection of the different patterns. We have evaluated the method using journalistic texts for a portion of the Federal Register corpus that contains a set of 313 documents (156,831 words). In the detection of pleonastic *it* pronouns a 88.7% **P** (568 out of 640) was obtained. Finally, it is very important to point out the high percentage of *it* pronouns in the test corpus that are pleonastic (32.9%). This fact demonstrates the importance of the correct detection of this kind of pronoun in any MT system.

Once the interlingua representation was obtained, the method proposed for pronominal anaphora translation into the target language was based on the treatment of number and gender discrepancies.

### 6.1 Pronominal Anaphora Translation into Spanish

In this experiment, the translation of English, third-person, personal pronouns into Spanish was evaluated.

We tested the method on the portions of the SemCor and MTI corpora used previously in the process of anaphora resolution. The training corpus was used for improving the number and gender rules. The remaining fragments of the corpora were reserved for test data.

We needed to know the semantic category (*person*, *animal*, or *object*) and the grammatical gender (*masculine* or *feminine*) of the pronoun's antecedent in order to apply the number and gender rules. In the SemCor corpus, the WordNet sense was used to identify the antecedent's semantic category. In the MTI corpus, due to the lack of semantic information, a set of heuristics was used to determine the antecedent's semantic category.

With regard to information about the antecedent's gender, an English–Spanish electronic dictionary was used since the POS tag does not usually provide gender and number information. The dictionary was incorporated into the system as a database. For each English word, the dictionary provides a translation into Spanish, and the word's gender and number in Spanish.

The number and gender rules were applied using this morphological and semantic information. We conducted a blind test over the entire test corpus, and the obtained results appear in Table 10.

| | Subject | Compl | Correct | Total | P(%) |
|---|---|---|---|---|---|
| SEMCOR | 197 | 47 | 229 | 254 | 90.2 |
| MTI | 239 | 231 | 353 | 470 | 75.1 |
| **TOTAL** | **436** | **288** | **582** | **724** | **80.4** |

Table 10: Translation of pronominal anaphora into Spanish, evaluation phase

The evaluation of this task was automatically carried out after the anaphoric annotation of each pronoun. This annotation included information about the antecedent and the translation into the target language of the anaphor. To do so, the human annotators translated the anaphors according to the criteria established by the morphological rules. For example, the pronoun *it* with subject function was translated into the Spanish pronoun *él* if its antecedent was of the *animal* type and *masculine*; on the other hand, if its antecedent was of the *object* type and *masculine*, it was translated into the Spanish pronoun *éste*; and so on. In the Spanish–English translation, the pronoun *él* with subject function was translated into the English pronoun *he* if its antecedent was a *person* type and *masculine*;

on the other hand, if its antecedent was an *object/animal* type and *masculine/feminine*, it was translated into the English pronoun *it*; and so on[20].

Table 10 shows the anaphoric pronouns of each corpus classified by grammatical function: *subject* and *complement* (direct or indirect object). The last three columns represent the number of pronouns successfully solved, the total number of solved pronouns, and the obtained precision, respectively. For instance, the SemCor corpus contains 197 pronouns with subject function and 47 complement pronouns. The precision obtained in this corpus was of 90.2% (229 out of 254).

**Discussion**. In the translation of English personal pronouns in the third person into Spanish, an overall precision of 80.4% (582 out of 724) was obtained. Specifically, 90.2% **P** and 75.1% **P** were obtained in the SemCor and MTI corpora, respectively.

From these results, we have extracted the following conclusions:

- In the SemCor corpus, all of the instances of the English pronouns *he*, *she*, *him*, and *her* were correctly translated into Spanish. There are two reasons for this:

  - The semantic roles of these pronouns were correctly identified in all of the cases.
  - These pronouns contain the necessary grammatical information (gender and number) that allows the correct translation into Spanish, independent of the antecedent proposed as a solution by the AGIR system.

  The errors in the translation of the pronouns *it*, *they*, and *them* were originated by the following different causes:

  - There were mistakes in the anaphora-resolution stage, that is, the antecedent proposed by the system was not the correct one (44.4% of the global mistakes). This caused an incorrect translation into Spanish mainly due to the fact that the proposed antecedent and the correct one had different grammatical genders.
  - There were mistakes in the identification of the semantic role of the pronouns that caused the application of an incorrect morphological rule (44.4%). These mistakes mainly originated in an incorrect process of clause splitting.
  - There were mistakes originated by the English–Spanish electronic dictionary (11.2%). Two circumstances could occur: (a) the word did not appear in the dictionary; and (b) the word's gender in the dictionary was different from the real word's gender, since the word had different meanings.

- In the MTI corpus, nearly all the pronouns were *it*, *they*, and *them* (96.2% of the total pronouns). The errors in the translation of these pronouns originated in the same causes as those in the SemCor corpus, although the percentages were different:

  - There were mistakes in the anaphora-resolution stage (22.9% of the mistakes).
  - There were mistakes in the identification of the pronouns' semantic role (62.9%).

20. In the automatic evaluation, a pronoun was considered as correctly translated when the pronoun proposed by the system was the same as that proposed by the human annotator. With this criterion, we evaluated the correct application of the corresponding morphological rule.

- There were mistakes that originated in the English–Spanish dictionary (14.2%). In this corpus, there were a large number of technical words that did not appear in the electronic dictionary.

- After analyzing the results, we observed that the precision of the SemCor corpus was approximately 15% higher than that obtained by the MTI corpus. The lower percentage obtained by the MTI corpus were the result of the corpus itself (most of the pronouns in this corpus are *it*, *they*, and *them*), and of the lack of semantic information.

In order to measure the efficiency of our proposal, we compared our system with one of the most representative MT systems of the moment: Systran. Systran was designed and built more than thirty years ago, and it is being continually modified in order to improve its translation quality. Moreover, it is easily accessible to Internet users through the service of MT on the web—BABELFISH[21]—which provides free translations between different languages. With regard to the problem of pronominal anaphora resolution and translation, Systran is one of the best MT systems studied (see Section 2) because, like our own system, it treats the problems of intersentential pronominal anaphora and Spanish zero pronouns on unrestricted texts after carrying out a partial parsing of the source text. As was mentioned in Section 2, a free trial of the commercial product SYSTRANLinks[22] was used to translate between the English and Spanish languages the evaluation corpora. The results appear in Table 11.

| | SYSTRANLinks | AGIR |
|---|---|---|
| SEMCOR | 75.4 | **82.5** |
| MTI | 58.1 | **69.3** |

Table 11: Translation of pronominal anaphora (complement pronouns only) into Spanish, SYSTRANLinks and AGIR

The evaluation of the SYSTRANLinks output was carried out by a human translator by hand. Pronouns judged as acceptable by the translator were considered correctly translated; otherwise, they were considered incorrectly translated.

Table 11 only shows the evaluation of English complement pronoun translation into Spanish because Systran did not translate all the subject pronouns into Spanish. By analyzing the Systran outputs of both corpora, we extracted the following conclusions:

- All the instances of the English pronouns *he* and *she* (always with subject function) were correctly translated into their Spanish equivalents *él* and *ella*.

- All the instances of the English pronouns *it* and *they* with subject function were omitted in Spanish—zero pronouns. These pronouns were not resolved in English, and subsequently were not translated into Spanish.

21. URL = `http://www.babelfish.altavista.com` (visited on 03/11/2002).
22. URL = `http://w4.systranlinks.com/config` (visited on 06/22/2002).

On the other hand, in our AGIR system, we have evaluated the correct application of the morphological rule to translate all source pronouns into target pronouns. A subsequent task must decide if the pronoun in the target language (a) must be generated as our system proposes, (b) must be substituted by another kind of pronoun (e.g., a possessive pronoun), or (c) must be eliminated (i.e., Spanish zero pronouns). Therefore, we have only taken into account the complement pronoun translation in order to make a fair comparison between the two systems.

As shown in Table 11, the precision obtained using AGIR is approximately 7–11% higher (depending on the corpus) than the one obtained using Systran. The errors in Systran originated in mistakes in the anaphora-resolution stage that caused incorrect translations, since the proposed antecedents and the correct ones have different grammatical gender. These errors can occur in intrasentential anaphors (as presented in Section 2) or in intersentential anaphors, as in the following example extracted from the corpora:

- (**E**) *[This information]$_i$* is only valid for Linux on the Intel platform. Much of *it$_i$* should be applicable to Linux on other processor architectures, but I have no first hand experience or information.

- (**S**) Esta información es solamente válida para Linux en la plataforma de Intel. Mucho de *él* debe ser aplicable a Linux en otras configuraciones del procesador, pero no tengo ninguna experiencia o información de primera mano.

This example shows an incorrect English–Spanish translation of the pronoun *it* done by Systran. In this case, the antecedent (*this information*, feminine) is in the previous sentence to the anaphor. It is incorrectly solved, and then it is incorrectly translated (the pronoun *él*—masculine—instead of the pronoun *ésta*—feminine).

### 6.2 Pronominal Anaphora Translation into English

In this experiment, the translation of Spanish, third-person, personal pronouns and zero pronouns (excluding reflexive pronouns) into English was evaluated. We tested the method on the portion of the LEXESP corpus that was previously used in the process of anaphora resolution.

We needed to know the semantic category and the grammatical gender of the pronoun's antecedent in order to apply the number and gender rules. In the LEXESP corpus, due to the lack of semantic information, a set of heuristics was used to determine the antecedent's semantic category. On the other hand, the information about the antecedent's gender was provided by the POS tag of the antecedent's head. We conducted a blind test over the entire test corpus, and the results appear in Table 12.

| | Subject | Compl | Correct | Total | P(%) |
|---|---|---|---|---|---|
| LEXESP | 630 | 145 | 657 | 775 | 84.8 |

Table 12: Translation of pronominal anaphora into English, evaluation phase

**Discussion**. In the translation of Spanish personal pronouns in the third person into English, an overall precision of 84.8% (657 out of 775) was obtained. From these results, we extracted the following conclusions:

- All the instances of the Spanish plural pronouns (*ellos*, *ellas*, *les*, *los*, *las*, and the zero pronouns in plural corresponding to the English pronouns *they* and *them*), were correctly translated into English. There are two reasons for this:
  - The semantic roles of these pronouns were correctly identified in all of the cases.
  - The equivalent English pronouns (*they* and *them*) lack gender information, that is, they are valid for masculine and feminine. Therefore, the antecedent's gender did not influence the translation of these pronouns.
- The errors occurred in the translation of the Spanish singular pronouns (*él*, *ella*, *le*, *lo*, *la*, and in zero pronouns in singular corresponding to the English pronouns *he*, *she*, *it*, *him*, and *her*). There were different causes for these errors:
  - There were mistakes in the anaphora-resolution stage (79.7% of the global mistakes), which caused an incorrect translation into Spanish, mainly due to the proposed antecedent and the correct one having different grammatical gender. Sometimes both had the same gender, but they had different semantic categories.
  - There were mistakes in the application of the heuristic used to identify the antecedent's semantic category (20.3%). This involved the application of an incorrect morphological rule.

Our proposal was compared with the SYSTRANLinks output. As shown in Table 13, the precision obtained by the AGIR system was approximately 28% higher than that obtained by Systran.

| | SYSTRANLinks | AGIR |
|---|---|---|
| LEXESP | 56.9 | **84.8** |

Table 13: Translation of pronominal anaphora into English, SYSTRANLinks and AGIR

The low results obtained in Systran are mainly the result of errors that occurred in the translation of Spanish zero pronouns. Specifically, out of 775 Spanish pronouns, 334 errors occurred, and 293 of them (87.7% of the global errors) originated in the translation of zero pronouns, whereas the remainder (12.3%) originated in the translation of the remaining not-omitted pronouns. The errors in the translation of zero pronouns mainly originated in their incorrect resolution.

## 7. Conclusion

In this paper we have evaluated the different tasks carried out in our MT approach (for Spanish and English languages) that allowed the correct pronominal anaphora translation into the target language. We have shown the importance of the resolution of anaphoric expressions in any MT system for correct translations into the target language, and how the main MT systems do not conveniently resolve this phenomenon.

Our approach, called AGIR, works on unrestricted texts to which partial-parsing techniques have been applied. After parsing and solving NLP problems, an interlingua representation of the entire text is obtained. This fact is one of the main advantages of our system since several problems (hardly solved by the majority of MT systems) can be treated and solved. These problems are the translation of intersentential anaphora, the detection of co-reference chains, and the translation of Spanish zero pronouns into English.

In the evaluation, we obtained a precision of 80.4% and 84.8% in the translation of Spanish and English pronominal anaphora, respectively. Previously, Spanish zero pronouns had been resolved (with a precision of 81.4%) and anaphoric personal pronouns had been resolved in English (with precisions of 86.6% and 76.8% in the SemCor corpus with semantic information and in the MTI corpus without it, respectively) and in Spanish (with a precision of 82.2%).

In addition, we carried out an exhaustive comparison with some well-known anaphora-resolution algorithms. Finally, we also compared pronoun translation with one of the most representative MT systems at the moment: Systran. In all of these comparisons, AGIR was shown to perform better.

A very important conclusion was extracted during the evaluation phase: the adding of semantic information improves the precision of the anaphora-resolution process considerably, and therefore the corresponding precision of the anaphora-translation process. Currently, the addition of this kind of information in the different stages of the AGIR system is being studied in order to improve the overall performance of the system.

The resolution and translation of new types of references, such as definite descriptions or anaphora originated by demonstrative pronouns, will be studied in the future. Moreover, the addition of new languages to the interlingua approach will be taken into account.

## Acknowledgments

The authors wish to thank Manuel Palomar for his helpful revisions of this paper; Ferrán Pla, Ruslan Mitkov, and Richard Evans for having contributed their corpora; and Rafael Muñoz, Maximiliano Saiz-Noeda, Patricio Martínez-Barco, and Juan Carlos Trujillo for their suggestions and willingness to help in any task related to this paper. We are also grateful for the helpful comments of the anonymous reviewers of several conference papers in which we presented our preliminary work.

This research has been supported by the Spanish Government, under projects TIC2000-0664-C02-02 and FIT-150500-2002-416.